# Separation Properties of Sets of Probability Measures

Fabio G. Cozman *
Escola Politécnica, Universidade de São Paulo
fgcozman@usp.br, http://www.cs.cmu.edu/~fgcozman

## Abstract

This paper analyzes independence concepts for sets of probability measures associated with directed acyclic graphs. The paper shows that *epistemic independence* and the standard Markov condition violate desirable separation properties. The adoption of a *contraction condition* leads to d-separation but still fails to guarantee a belief separation property. To overcome this unsatisfactory situation, a *strong Markov condition* is proposed, based on epistemic independence. The main result is that the strong Markov condition leads to *strong independence* and does enforce separation properties; this result implies that (1) separation properties of Bayesian networks do extend to epistemic independence and sets of probability measures, and (2) strong independence has a clear justification based on epistemic independence and the strong Markov condition.

## 1 Introduction

Sets of probability measures, called *credal sets* by Levi [17], are quite flexible representations for uncertainty, employed in several theories [3, 12, 14, 15, 17, 18, 20, 21, 23]. Credal sets can be used to represent the opinions of a group of experts, to model the imprecision and incompleteness of realistic decision making, and to investigate the effect of perturbations in standard probabilistic models.

This paper focuses on credal sets that are represented by directed acyclic graphs. Such structures are called *credal networks*[1] and have been investigated by several authors [1, 4, 11, 22]. The idea is to associate a local credal set with each node in a directed acyclic graph, mimicking the theory of Bayesian networks [19]. Figure 1 shows a simple credal network; note that all nodes are associated with probability intervals. Sections 2 and 3 briefly review the main ideas connected with credal sets and credal networks.

There are two difficulties with credal networks. First, there are several concepts of independence for credal sets. Second, there are several ways to combine locally defined credal sets in a credal network, i.e., several *extensions* can be defined for a given network. Consequently, it is hard to tell exactly what a credal network represents. Take the network in Figure 1: Is it representing a credal set where $W$ and $Y$ are *epistemically* independent given $X$, or a credal set where $W$ and $Y$ are *strongly* independent given $X$? Is it representing the largest credal set for a given concept of independence, or a credal set constructed in some special way?

The objective of this paper is to present a theory of credal networks that overcomes the difficulties described in the previous paragraph. The idea is to analyze concepts of independence and methods of extension through their separation properties. Sections 4 and 5 present current approaches to credal networks and analyze their weaknesses. Based on this analysis, a new condition, called the *strong Markov condition*, is presented (Section 6). The strong Markov condition can alone organize the theory of credal sets: it implies that a small number of intuitive elements leads to a unique, computationally tractable type of extension for credal networks, with pleasant separation properties.

The argument leading to the strong Markov condition can be summarized as follows. There are two important concepts of independence for credal sets: *strong* and *epistemic* independence (Section 2). Strong independence leads to *strong extensions*, which inherit

---

* The author was partially supported by a grant from CNPq, Brazil.

[1] The author has employed the term *Quasi-Bayesian* networks [7, 8]; the term *credal networks*, proposed by Zaffalon [24], seems more appropriate.

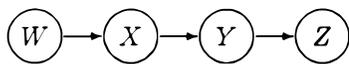

$0.2 \leq p(w) \leq 0.3$
$0.1 \leq p(x|w) \leq 0.2 \quad 0.8 \leq p(x|w^c) \leq 0.9$
$0.4 \leq p(y|x) \leq 0.5 \quad 0.5 \leq p(y|x^c) \leq 0.6$
$0.7 \leq p(z|y) \leq 0.8 \quad 0.1 \leq p(z|y^c) \leq 0.2$

Figure 1: Credal network with four binary variables (superscript $c$ indicates negation).

several nice features of Bayesian networks, including computational simplicity (Section 4). But strong independence is mostly a mathematical generalization of standard stochastic independence without a clear, direct justification. Section 5 moves the focus from strong independence to epistemic independence. Section 5 investigates the separation properties of extensions based on epistemic independence and the standard Markov condition. The rationale for this investigation is that the separation properties of standard stochastic independence are fundamental in the theory of Bayesian networks. In fact, separation properties are not just implied by stochastic independence, separation properties *validate* the concept of stochastic independence. Section 5 presents a number of novel examples to demonstrate that epistemic independence and the standard Markov condition violate desirable separation properties; the examples also indicate that epistemic independence leads to extensions that are intractable in practice. Given the weaknesses of the extensions analyzed in Sections 4 and 5, we are led to ask whether there is some intuitive condition on a credal network that (1) is based on epistemic independence, and (2) produces a suitable, tractable type of extension. The strong Markov condition is such a condition, as argued in Section 7.

## 2 Credal sets

This section contains basic definitions and notation used in the paper. For discrete random variables $X$ and $Y$, $p(X)$ denotes the probability density of $X$, $p(X|y)$ denotes the conditional density of $X$ given the event $\{Y = y\}$, $E_p[f(X)]$ denotes the expectation of function $f(X)$ with respect to $p(X)$, and $E_p[f(X)|y]$ denotes the expectation of function $f(X)$ with respect to $p(X|y)$.

A credal set $K$ is a collection of probability measures. A credal set defined by a collection of densities $p(X)$ is denoted by $K(X)$. Given a credal set $K(X)$ and a function $f(X)$, the *lower* and *upper* expectations of $f(X)$ are defined respectively as

$\underline{E}[f(X)] = \min_{p(X) \in K(X)} E_p[f(X)]$ and $\overline{E}[f(X)] = \max_{p(X) \in K(X)} E_p[f(X)]$. A credal set defines a unique lower expectation for every bounded function. Similarly, the *lower probability* and the *upper probability* of event $A$ are defined respectively as $\underline{P}(A) = \min_{p(X) \in K(X)} P(A)$ and $\overline{P}(A) = \max_{p(X) \in K(X)} P(A)$.

A set of probability measures and its convex hull produce the same lower and upper expectations. The notation $\otimes K$ indicates the convex hull of a set of measures $K$ and the notation $\text{ext}K$ indicates the extreme points of a credal set $K$.

Inference is performed by applying Bayes rule to each measure in a credal set; the posterior credal set is the union of all posterior probability measures [14]. A *conditional* credal set $K(X|y)$ contains densities $p(X|y)$ for random variables $X$ and $Y$.

There are several concepts of independence that can be applied to credal sets [6, 10]. To understand them, consider the two definitions of stochastic independence in probability theory. Variables $X$ and $Y$ are stochastically independent either if $p(X|y) = p(X)$ for any value $y$, or if $p(X,Y) = p(X)p(Y)$ (appropriate positivity conditions may be required). Several authors have tried to adapt both methods to the theory of credal sets using various multiplication operators and conditioning rules. The next subsections describe concepts that can be cast in the language of credal sets.

### 2.1 Epistemic independence

The idea of epistemic independence [23, Chap. 9] is to start from an asymmetric irrelevance relation and then define independence from irrelevance.

**Definition 1** *Variable $Y$ is epistemically irrelevant to $X$ given $Z$, denoted by $(Y \text{ EIR } X \mid Z)$, if $K(X|z)$ and $K(X|y,z)$ have the same convex hull for all possible values of $Y$ and $Z$.*

In terms of lower expectations, $(Y \text{ EIR } X \mid Z)$ iff $\underline{E}[f(X)|y,z] = \underline{E}[f(X)|z]$ for any function $f(X)$ and all possible values of $Y$ and $Z$. Epistemic irrelevance gets an intuitive significance when we think that lower and upper expectations are the practical consequences of beliefs, as they are used for decision-making and inference. Irrelevance of $Y$ to $X$ means that the practical consequences of our beliefs about $X$ are not affected by knowledge of $Y$. If we take two credal sets to be *equivalent* when their convex hulls agree, we have that $(Y \text{ EIR } X \mid Z)$ iff $K(X|z)$ and $K(X|y,z)$ are equivalent for all possible values of $Y$ and $Z$. A more stringent definition would be to impose *equality* between $K(X|z)$ and $K(X|y,z)$ for all possible values of $Y$ and $Z$. If $K(X|z)$ and $K(X|y,z)$ are non-convex, then equality and equivalence have different meanings.

Epistemic independence is just a symmetrization of epistemic irrelevance.

**Definition 2** *Variables $X$ and $Y$ are* epistemically independent *given $Z$, denoted by $(X \text{ EIN } Y \mid Z)$, if $(X \text{ EIR } Y \mid Z)$ and $(Y \text{ EIR } X \mid Z)$.*

### 2.2 Strong independence

Consider the following concept:

**Definition 3** *Variables $X$ and $Y$ are* strongly independent *when every extreme point of $K(X,Y)$ satisfies standard stochastic independence of $X$ and $Y$.*

This concept is called *independence in the selection* by Couso et al [6]; they employ *strong independence* when the joint credal set is the largest credal set that satisfies Definition 3. Here we prefer to use strong independence and take the largest credal set as the *strong extension* (Section 4).

We can even ask that *every* probability density in a credal set factorizes according to stochastic independence. In this case the resulting credal set is nonconvex [16]. Note that from a behavioral point of view, insisting on stochastic independence for all probabilities in a credal set is excessive because it suggests that a credal set and its convex hull are different. But two individuals with the exact same lower and upper expectations should have equivalent beliefs, at least from a behavioral perspective. On the other hand, adopting convexity significantly diminishes the justification for stochastic independence. Even if all extreme points of a convex credal set display independence between two variables, there may be densities in the credal set that display dependence between the variables [2, 5]. To accommodate these conflicting views, the definition of strong independence does not require (nor prohibits) convexity.

Strong independence is reasonable in a "sensitivity analysis" interpretation of credal sets: a credal set is then viewed as a set containing a "true" probability, and not as a representation of uncertainty in itself [23]. Strong independence is also reasonable when a collection of experts, specifying a credal set, agrees that every lower expectation must be computed with respect to a density that displays stochastic independence.

Despite the intuitive appeal of strong independence, there is no known direct justification for it — a justification that does not employ stochastic independence as a starting point. Definitions of strong independence usually assume stochastic independence and standard probability — a most undesirable situation, as the theory of credal sets purports to offer a more basic approach to uncertainty, an approach that contains standard probability theory as a limiting case. Strong independence stands as a mathematically-minded generalization of the factorization property of standard stochastic independence.

One concept of independence that might characterize strong independence through lower expectations is Kuznetsov's independence. Variables $X$ and $Y$ are (Kuznetsov-)independent given $Z$ if, for any functions $f(X)$ and $g(Y)$ and all values of $Z$,

$$\overline{E}[f(X)g(Y)|z] = \overline{E}[f(X)|z] \times \overline{E}[g(Y)|z], \quad (1)$$

where $\overline{\underline{E}}[f(X)]$ indicates $[\underline{E}[f(X)], \overline{E}[f(X)]]$ and $\times$ indicates *interval* multiplication. It is an open question whether this concept is equivalent to some form of strong independence; even if it were, Expression (1) cannot be easily justified except as a mathematical generalization of standard stochastic independence.[2]

## 3 Graphical models

Multivariate statistical models can be elegantly represented by graphical models, as demonstrated by the theory of Bayesian networks [19]. A Bayesian network is composed of a directed acyclic graph and a collection of variables **X**. Each node in the graph is associated with a variable $X_i$ and with a conditional density $p(X_i|\text{pa}(X_i))$. Bayesian networks satisfy the Markov condition: Every variable is independent of its nondescendants non-parents given its parents. The Markov condition implies that every Bayesian network represents a unique joint probability

$$p(\mathbf{X}) = \prod_i p(X_i|\text{pa}(X_i)). \quad (2)$$

Credal networks are structures that represent joint credal sets through directed acyclic graphs; for this purpose, define [8]:

**Definition 4** *A* locally defined credal network *for a collection of variables* **X** *is a directed acyclic graph where every node is associated with a variable $X_i$ and with a* local credal set $K(X_i|\text{pa}(X_i))$.

A joint credal set $K(\mathbf{X})$ that satisfies all constraints in a credal network is an *extension* of the network. Note that Definition 4 does not generate a *unique* extension for a given credal network [8]. The flexibility of Definition 4 seems justified as a starting point, as it would be unduly restrictive to select an extension, among the many possible extensions, as the "correct" one. The next sections look at strategies that automatically specify the independence judgements in credal networks; the analysis will lead to the strong Markov condition (Section 6).

---

[2]Unfortunately, Kuznetsov died in an accident at the beginning of 1998, and little is known about his concept of independence.

## 4 Strong extensions

What happens if every variable in a credal network is strongly independent of its nondescendants non-parents given its parents? We can ask for the *largest* possible credal set that satisfies this type of Markov condition; call this credal set the *strong extension* of the network.

Strong extensions have a rather simple form. As proved in Appendix A:

**Theorem 1** *The strong extension of a locally defined credal network with convex local credal sets is the convex hull of all probability densities that satisfy the Markov condition on the network when each conditional probability $p(X_i|\text{pa}(X_i))$, for each value of $\text{pa}(X_i)$, is selected from $K(X_i|\text{pa}(X_i))$:*

$$K(\mathbf{X}) = \otimes \prod_i p(X_i|\text{pa}(X_i)). \quad (3)$$

This result is apparently assumed, or at least known informally, in most of the literature that associates credal sets with directed acyclic graphs [4, 7, 11, 22]. The convexity of the local credal sets $K(X_i|\text{pa}(X_i))$ is quite a reasonable assumption that is usually taken for granted. Most of the literature starts from Expression (3) and argues that this expression is the "correct" extension of a credal network, rather than obtaining it as a consequence of strong independence.[3]

Strong extensions have an intuitive similarity with standard Bayesian networks, and they satisfy one of the most important properties of Bayesian networks (as proved in Appendix A):

**Theorem 2** *Given a credal network where all combinations of variables have positive probability, every d-separation relation in the network corresponds to a strong independence in the strong extension of the network.*

Actually, d-separation relations even imply *epistemic independence* in strong extensions [8], so the similarity to Bayesian networks is quite remarkable.

---

[3]The author has employed the term "type-1 extension" to refer to Expression (3) [7, 8].

On the negative side, strong extensions are just as hard to justify as strong independence. They make sense in a "sensitivity analysis" interpretation of credal sets, but they can only have as compelling a justification as strong independence.

## 5 Extensions based on epistemic independence

Suppose we adopt epistemic independence as a solid and compelling concept of independence, and we adopt the standard Markov condition: Every variable in a credal network is epistemically independent of its non-descendants non-parents given its parents. What separation properties are valid in this model? The following example shows that d-separation does not imply epistemic independence.

**Example 1** A group of experts models a domain by the credal network in Figure 1. The experts reach a preliminary model by adopting all extreme points of the strong extension that satisfy either $(p(z|y), p(z|y^c)) = (0.7, 0.2)$ or $(p(z|y), p(z|y^c)) = (0.8, 0.1)$. Denote the resulting extension by $K'(Z, Y, X, W)$ (this extension has 64 extreme points). Table 1 shows two extreme points of $K'(Z, Y, X, W)$ (denoted by $p'_1$, $p'_2$). After additional discussion, the experts conclude that an additional probability measure $p^*$ must be added to the extension (also shown in Table 1). The experts agree on the extension $K''(Z, Y, X, W) = \{p^* \cup K'(Z, Y, X, W)\}$, because this credal set satisfies (tightly) all probability bounds and also satisfies the Markov condition for epistemic independence: $(W \text{ EIN } Y \mid X)$ and $((W, X) \text{ EIN } Z \mid Y)$.

Unfortunately, d-separation does not imply epistemic independence in the extension $K''(Z, Y, X, W)$. Note that $Z$ and $W$ are d-separated by $X$, but it is *not* true that $(W \text{ EIN } Z \mid X)$. For example, $\underline{p}(z|x) = 8501/22707 < 19/50 = \underline{p}(z|x, w)$. ◊

Of course, if d-separation were to be maintained at all costs, we could simply impose it. We would say that, in any extension of a credal network, d-separation *must* imply epistemic independence. If d-separation seems too convoluted to be adopted outright, we can optionally adopt the following condition:

**Definition 5 (Contraction condition)** *Suppose $\mathbf{W}$, $\mathbf{X}$, $\mathbf{Y}$, and $\mathbf{Z}$ are collections of variables in a credal network and $\mathbf{X}$, $\mathbf{Y}$ and $\mathbf{Z}$ are non-descendants of $\mathbf{W}$. If $\mathbf{Y}$ is epistemically irrelevant to $\mathbf{X}$ given $\mathbf{Z}$, and $\mathbf{Y}$ is epistemicallly irrelevant to $\mathbf{W}$ given $(\mathbf{X}, \mathbf{Z})$, then $\mathbf{Y}$ is epistemically irrelevant to $(\mathbf{W}, \mathbf{X})$ given $\mathbf{Z}$.*

|       | $zyxw$ | $zyxw^c$ | $zyx^cw$ | $zyx^cw^c$ | $zy^cxw$ | $zy^cxw^c$ | $zy^cx^cw$ | $zy^cx^cw^c$ |
|-------|--------|----------|----------|------------|----------|------------|------------|--------------|
| $p'_1$ | 7/1250 | 112/625 | 63/1000 | 7/125 | 3/1250 | 48/625 | 9/500 | 2/125 |
| $p'_2$ | 4/625 | 128/625 | 54/625 | 48/625 | 3/2500 | 24/625 | 9/1250 | 4/625 |
| $p^*$ | 211/17000 | 609/3400 | 3717/34000 | 203/6800 | 81/17000 | 1827/34000 | 177/8500 | 203/34000 |

|       | $z^cyxw$ | $z^cyxw^c$ | $z^cyx^cw$ | $z^cyx^cw^c$ | $z^cy^cxw$ | $z^cy^cxw^c$ | $z^cy^cx^cw$ | $z^cy^cx^cw^c$ |
|-------|----------|------------|------------|--------------|------------|--------------|--------------|----------------|
| $p'_1$ | 3/1250 | 48/625 | 27/1000 | 3/125 | 6/625 | 192/625 | 9/125 | 8/125 |
| $p'_2$ | 1/625 | 32/625 | 27/1250 | 12/625 | 27/2500 | 216/625 | 81/1250 | 36/625 |
| $p^*$ | 59/17000 | 1239/17000 | 1593/34000 | 413/34000 | 81/4250 | 5481/17000 | 177/2125 | 203/8500 |

Table 1: Probability measures for Example 1.

The Markov and the contraction conditions lead to d-separation properties for epistemic independence, when we require *equivalence* of credal sets for epistemic independence (Definition 1). In this case, the conditions allow a duplication of the d-separation proof for Bayesian networks [13]. The contraction condition is necessary because epistemic independence does *not* satisfy precisely the graphoid property of contraction employed in Verma, Geiger and Pearl's proof of d-separation [9].

Note that the contraction condition is not satisfied in Example 1. Note also that if we require *equality* between credal sets for epistemic independence, other graphoid properties may fail and the contraction condition may not suffice for d-separation.

The contraction condition may itself look too convoluted and a bit excessive, for the contraction and Markov conditions guarantee d-separation for *all* extensions of a network. Instead, we may ask for the separation properties of particular extensions. Take the largest credal set that complies with (1) the constraints that define the local credal sets in a credal network; and with (2) the Markov condition for epistemic independence. Call this credal set the *independent natural extension* of the credal network. In general, a *natural extension* is simply the largest credal set satisfying a collection of constraints [23].

Independent natural extensions are quite intuitive, but little is known about them — except that the Markov condition generates far too many constraints. The independent natural extension of the network in Figure 1 has more than 6 million extreme points! Whether or not d-separation implies epistemic independence in independent natural extensions is an important open question with no obvious answer. Finding a counterexample for d-separation relations, if there is one, is quite an undertaking, as even simple networks produce quite complex independent natural extensions.

On top of the complexity issues, independent natural extensions do not satisfy a property that we might call *belief separation*, as illustrated in the next example.

**Example 2** Consider two binary variables $X$ and $Y$, such that $(X \text{ EIN } Y)$ and $p(X) \in [2/5, 1/2]$, $p(Y) \in [2/5, 1/2]$ [23, Sect. 9.3.4]. This can be represented by a trivial credal network with two unconnected nodes. The strong extension of this network has four extreme points (vectors $[p(x,y), p(x,y^c), p(x^c,y), p(x^c,y^c)]$):

$$[1/4, 1/4, 1/4, 1/4], [4/25, 6/25, 6/25, 9/25],$$

$$[1/5, 1/5, 3/10, 3/10], [1/5, 3/10, 1/5, 3/10].$$

The independent natural extension has six extreme points: the four points of the strong extension and $[2/9, 2/9, 2/9, 1/3]$, $[2/11, 3/11, 3/11, 3/11]$. An expert then communicates that $p(y)$ is exactly 2/5. To simplify our calculations, we simply go through the six densities in the independent natural extension, changing $p(y)$ to 2/5. As $X$ and $Y$ are independent, we reason that there is no need to change the conditional densities $p(X|Y)$ and recompute the independent natural extension from scratch. Doing so, we obtain four different extreme points:

$$[4/25, 6/25, 6/25, 9/25], [1/5, 3/10, 1/5, 3/10],$$

$$[1/5, 6/25, 1/5, 9/25], [4/25, 3/10, 6/25, 3/10].$$

The problem is that $X$ and $Y$ are *not* epistemically independent in this new extension; for example, $p(y) = 2/5$ but $p(y|x) \in [2/5, 4/7]$. ◇

A similar situation can be informally described as follows. Take three variables, $X$ as {rain, no rain}, $Y$ as {airport on, airport of}, $W$ as {good, bad} for the weather at the remote location. Suppose we have the structure of Figure 1, including the relation $(Y \text{ EIN } W \mid X)$. The independent natural extension of this network introduces a form of "linkage" between the variables, because conditional densities for $X$ and $Y$ may be tied to particular marginal densities for $W$. If we find that an extreme point (with linkage) is inadequate, we cannot change $p(W)$ or $p(X|W)$ or $p(Y|X)$ for that extreme point in isolation. Independent manipulation of credal sets is not possible with epistemic independence and the standard Markov condition.

## 6 The strong Markov condition

The previous sections made a detailed analysis of independence concepts and methods of extension in credal networks, and the result is far from satisfactory. Strong extensions are quite manageable computationally and satisfy d-separation, but there is no justification for strong extension as long as strong independence remains only a mathematical generalization of standard stochastic independence. Epistemic independence has a much better justification, but does not guarantee desirable separation properties.

The key idea here is to recognize that the standard Markov condition, and not epistemic independence, is too weak. In essence, the standard Markov condition demands that beliefs about a situation should not be affected by the past once we know everything that causes that situation. But in fact we can say more: our beliefs about a situation should not be affected by the past, *and by changes in our beliefs about the past*, once we know everything that causes that situation.

To obtain an improvement on the standard Markov condition, consider the following interpretation for the expression "changes in our beliefs." Suppose a joint credal set $K(X,Y)$ is given. Every extreme point of $K(X,Y)$ can be written as $p(X|Y)p(Y)$. Select an extreme point $p(X,Y)$, remove it from the collection of extreme points of the credal set and modify $p(X,Y)$: replace $p(Y)$ by an arbitrary $p'(Y)$. Then insert $p(X|Y)p'(Y)$ into the collection of extreme points and take the convex hull of all densities there. Call any such modification of $K(X,Y)$ a *belief change with respect to* $Y$.

This rationale leads to the following condition:

**Definition 6 (Strong Markov condition)** *Every variable $X_i$ is epistemically independent of its non-descendants non-parents given its parents, regardless of any sequence of belief changes with respect to the nondescendants of $X_i$.*

Denote the nondescendants non-parents of $X_i$ by $\text{nD}(X_i)$. Intuitively, the strong Markov condition requires that $(X_i \text{ EIR } \text{nD}(X_i) \mid \text{pa}(X_i))$ regardless of $K(\text{pa}(X_i), \text{nD}(X_i))$.

Despite the intuitive character of the strong Markov condition, it is not clear whether this condition is useful or not — i.e., whether or not it can lead to interesting separation properties and important computational simplifications. The main result of the paper, expressed by the next theorem, is that the strong Markov condition implies the standard Markov condition with strong independence (the proof is in Appendix A). The impact of this result is discussed in Section 7.

**Theorem 3** *In a credal network with convex local credal sets, the strong Markov condition holds if and only if every variable $X_i$ is strongly independent of its nondescendants non-parents given its parents.*

To illustrate the consequences of Theorem 3, consider a quite common model in statistics: an experiment is independently repeated several times. Here we must differentiate between the assumption that the experiment is independently repeated and the assumption that the repeated experiments are in fact *exchangeable* — that is, the exact same probabilistic model should represent every repetition of the experiment. Here we assume only independence; Walley discusses at length the concept of exchangeability and its implications in the theory of credal sets [23, Chap. 9].

**Example 3** Take a collection of variables $X_j$, $j \in \{1,\ldots,n\}$, where every variable is independent of all others and all variables can be modeled by the same equivalent credal sets. Suppose the marginal convex credal set $K(X_j)$ is given for all variables $X_j$. If we assume the strong Markov condition, then the natural extension of this credal network is the strong extension $\otimes \prod_{j=1}^{n} p_j(X_j)$, for every combination of $p_j(X_j) \in \text{ext}K(X_j)$. ◇

## 7 Discussion

The main contribution of this paper is the strong Markov condition. To understand the place of the strong Markov condition in the theory of credal networks, note that Theorems 1 and 3 demonstrate the equivalence of:

1. Strong Markov condition (with epistemic independence) plus natural extension.

2. Standard Markov condition (with strong independence) plus natural extension.

3. Strong extension.

So, epistemic independence does satisfy the "belief separation" property through the strong Markov condition is adopted. Note also that, in strong extensions, d-separation implies *epistemic* independence [7], so epistemic independence and natural extension do display the same d-separation relations displayed by strong extensions and standard Bayesian networks. This argument proves the "soundness" of d-separation (that every d-separation relation produces an independence relation). The "completeness" of d-separation (there is always a joint credal set where variables that are not d-separated are dependent) comes from the fact that every standard Bayesian network is a credal network.

Consequently, epistemic independence and natural extension do display the essential separation properties of standard Bayesian networks, provided we accept the strong Markov condition. Using the strong Markov condition, we obtain the computational simplicity associated with the strong extension. As a by-product, we obtain a clear justification for strong independence; namely, that strong independence is a consequence of epistemic independence and the strong Markov condition.

As discussed in Section 2.2, the definition of strong independence has two drawbacks: it deals with individual probabilities in a credal set, and, rather unfortunately, it requires an understanding of stochastic independence. The strong Markov condition solves the latter, more serious, problem as it does not require any concept of independence for individual probabilities. The strong Markov condition still relies on individual probabilities. But note that individual probabilities appear in a progression that is quite satisfactory in the theory of credal networks described here. The theory first adopts epistemic independence, then deals with multivariate and graphical representations, and only then introduces individual probabilities to define belief changes. Future work should attempt to improve the results here by defining belief changes in terms of credal sets.

In short, if we accept epistemic independence as *the* definition of independence and natural extension as *the* method of extension, we have a pleasant relationship:

$$\text{Strong Markov condition} \Leftrightarrow \text{Strong extension.} \quad (4)$$

Note that Expression (4) closely resembles the theory of Bayesian networks, where
$$\text{standard Markov condition} \Leftrightarrow \text{Expression (2)}.$$

So we can build a powerful theory of credal networks starting from five elements: directed acyclic graphs and local credal sets, epistemic independence, the strong Markov condition and natural extension. Ultimately, Expression (4) is the reason why the strong Markov condition can be the foundation of a solid and appealing theory of credal networks.

## A  Proofs

**Theorem 1:** The strong extension must be equal or larger than $\otimes \prod_i p(X_i|\text{pa}(X_i))$, because this credal set satisfies the standard Markov condition for strong independence. But this set is the largest possible extension, because any larger set will have some $p(X_i|\text{pa}(X_i), \text{nD}(X_i))$ as a non-constant function of $\text{nD}(X_i)$ — violating a strong independence relation required by the standard Markov condition.  ◊

**Theorem 2:** Suppose **X** and **Y** are d-separated by **Z**; then every extreme point of the strong extension satisfies stochastic independence of **X** and **Y** given **Z** and consequently **X** and **Y** are strongly independent given **Z**.  ◊

**Theorem 3:** ⇐ If for every variable $X_i$, $X_i$ and $\text{nD}(X_i)$ are strongly independent given $\text{pa}(X_i)$, then the strong Markov condition is satisfied. To see that, note that as any extreme point of an extension must satisfy $p(X_i|\text{pa}(X_i), \text{nD}(X_i)) = p(X_i|\text{pa}(X_i))$, any extreme point must satisfy $(X_i \text{ SIN } \text{nD}(X_i) \mid \text{pa}(X_i))$ regardless of $K(\text{nD}(X_i))$.

⇒ Suppose the strong Markov condition holds for a given extension $K(\mathbf{X})$. Denote $D_i = \text{nD}(X_i)$ and $A_i = \text{pa}(X_i)$. Take a particular variable $X_i$, and select $K'(A_i, D_i)$ to be a singleton set containing an arbitrary everywhere positive density $p'(A_i, D_i)$. This can be done arbitrarily because the strong Markov condition is valid for any sequence of belief changes, and a sequence can always be built to produce an arbitrary $p'(A_i, D_i)$. Note that $K'(D_i|a_i)$ is a singleton set for any value of $A_i$, with density $p'(D_i|a_i)$. Consider the extreme points of $K(X_i, A_i, D_i)$ (the marginal credal set of extension $K(\mathbf{X})$) and work by reductio. Suppose that for one of the extreme points of $K(X_i, A_i, D_i)$, denoted by $p^*(X_i, A_i, D_i)$, the density $p^*(X_i|a_i, d_i)$ is a non-constant function of $D_i$ for some value of $A_i$. If this is true, there must be some value $(x_i, a_i, d_i)$ of $(X_i, A_i, D_i)$ such that $p^*(x_i|a_i, d_i) < p^*(x_i|a_i)$. Now consider the unique extension of $p'(A_i|D_i)$ and $K(X_i|a_i, d_i)$ (the conditional credal set of extension $K(\mathbf{X})$); this unique extension is equal to $p'(A_i|D_i) \times K(X_i|A_i, D_i)$, where × indicates elementwise scalar multiplication. Note that the density $p'(A_i, D_i) p^*(X_i|A_i, D_i)$ must belong to this unique extension. For some value $(x_i, a_i, d_i)$:

$$p'(d_i|a_i)\, p^*(x_i|a_i, d_i) < p^*(x_i|a_i) p'(d_i|a_i),$$

$$\Rightarrow p'(d_i|a_i)\, p^*(x_i|a_i, d_i)/p^*(x_i|a_i) < p'(d_i|a_i),$$

$$\Rightarrow p^*(d_i|x_i, a_i) < p'(d_i|a_i).$$

Then $\underline{p}(d_i|x_i, a_i) < \underline{p}(a_i|d_i)$, contradicting the strong Markov condition. So every extreme point of $K(X_i, A_i, D_i)$ must have its conditional density $p(X_i|A_i, D_i)$ as a constant function of $D_i$, and consequently every variable $X_i$ is strongly independent of $D_i$ given $A_i$.  ◊

## Acknowledgements

I greatly benefited from joint work with Peter Walley on graphoid properties; I learned about Kuznetsov's independence from him. Thanks to David Avis for the lrs package, used in Example 1 (lrs package is currently available at ftp://mutt.cs.mcgill.ca/pub/C/lrs.html).